\documentclass{article}

\usepackage{PRIMEarxiv}

\usepackage{fancyhdr}       

\usepackage{graphicx}%
\usepackage{multirow}%
\usepackage{amsmath,amssymb,amsfonts}%
\usepackage{amsthm}%
\usepackage{mathrsfs}%
\usepackage[figuresright]{rotating}%
\usepackage[title]{appendix}%
\usepackage{xcolor}%
\usepackage{textcomp}%
\usepackage{manyfoot}%
\usepackage{booktabs}%
\usepackage{vruler}%
\usepackage{setspace}%
\usepackage{listings}%

\usepackage{latexsym}
\usepackage{caption}
\usepackage{subcaption}
\usepackage{verbatim}
\usepackage[linesnumbered,algoruled]{algorithm2e}
\usepackage{tikz, pgfplots}
\usetikzlibrary{positioning, decorations.pathmorphing}

\tikzset{snake it/.style={decorate, decoration={snake, amplitude=0.5mm}}}

\usepackage[authoryear]{natbib}

\newcommand{\ts}{\tilde{s}}

\newtheorem{theorem}{Theorem}
\newtheorem{prop}{Proposition}%
\newtheorem{property}{Property}%
\newtheorem{lemma}{Lemma}%
\newtheorem{corollary}{Corollary}%

\title{Normalizing flow sampling \\with Langevin dynamics in the latent space
}

\author{
  Florentin Coeurdoux, Nicolas Dobigeon\thanks{This work was supported by the Artificial Natural Intelligence Toulouse Institute (ANITI, ANR-19-PI3A-0004).}  \\
  IRIT/INP-ENSEEIHT \\
  University of Toulouse \\
  Toulouse, France\\
  \texttt{\{Florentin.Coeurdoux, Nicolas.Dobigeon\}@irit.fr} \\
   \And
  Pierre Chainais\thanks{This work was supported by the AI Sherlock Chair (ANR-20-CHIA-0031-01), the ULNE national future investment programme (ANR-16-IDEX-0004) and the Hauts-de-France Region.}  \\
  CNRS, Centrale Lille \\
  University of Lille \\
  Lille, France\\
  \texttt{Pierre.Chainais@centralelille.fr}
}

\begin{document}
\maketitle

\begin{abstract}
Normalizing flows (NF) use a continuous generator to map a simple latent (e.g. Gaussian) distribution, towards an empirical target distribution associated with a training data set. Once trained by minimizing a variational objective, the learnt map provides an approximate generative model of the target distribution. Since standard NF implement differentiable maps, they may suffer from pathological behaviors when targeting complex distributions. For instance, such problems may appear for distributions on multi-component topologies or characterized by multiple modes with high probability regions separated by very unlikely areas. A typical symptom is the explosion of the Jacobian norm of the transformation in very low probability areas. This paper proposes to overcome this issue thanks to a new Markov chain Monte Carlo algorithm to sample from the target distribution in the latent domain before transporting it back to the target domain. The approach relies on a Metropolis adjusted Langevin algorithm (MALA) whose dynamics explicitly exploits the Jacobian of the transformation. Contrary to alternative approaches, the proposed strategy preserves the tractability of the likelihood and it does not require a specific training. Notably, it can be straightforwardly used with any pre-trained NF network, regardless of the architecture.
Experiments conducted on synthetic and high-dimensional real data sets illustrate the efficiency of the method.
\end{abstract}


\section{Introduction}

Normalizing flows (NFs) are known to be a very efficient generative model to approximate probability distributions in an unsupervised setting. For example, {\em Glow} \citep{kingma2018glow} is able to generate very realistic human faces, competing with state-of-the-art algorithms of variational inference \citep{papamakarios2021normalizing}. Despite some early theoretical results about their stability \citep{nalisnick2018deep} or their approximation and asymptotic properties \citep{behrmann2019invertibility}, their training remains challenging in the most general cases. 
Their capacity is limited by intrinsic architectural constraints, resulting in a variational mismatch between the target distribution and the actually learnt distribution.
In particular \citet{cornish2020relaxing} pointed out the capital issue of target distributions with disconnected support featuring several components. Since NFs provide a continuous differentiable change of variable, they are not able to deal with such distributions when using a monomodal (e.g., Gaussian) latent distribution. 
Even targeting multimodal distributions featuring high probability regions separated by very unlikely areas remains problematic. 
%
%
The trained NF is a continuous differentiable transformation so that the transport of latent samples to the target space may overcharge low probability areas with (undesired) samples. These out-of-distribution samples will correspond to smooth transitions between different modes, which leads to out-of-distribution samples, as discussed by \citet{cornish2020relaxing}.

\begin{figure}
    \centering
    \begin{tabular}{ccc}
    \includegraphics[width=0.3\linewidth]{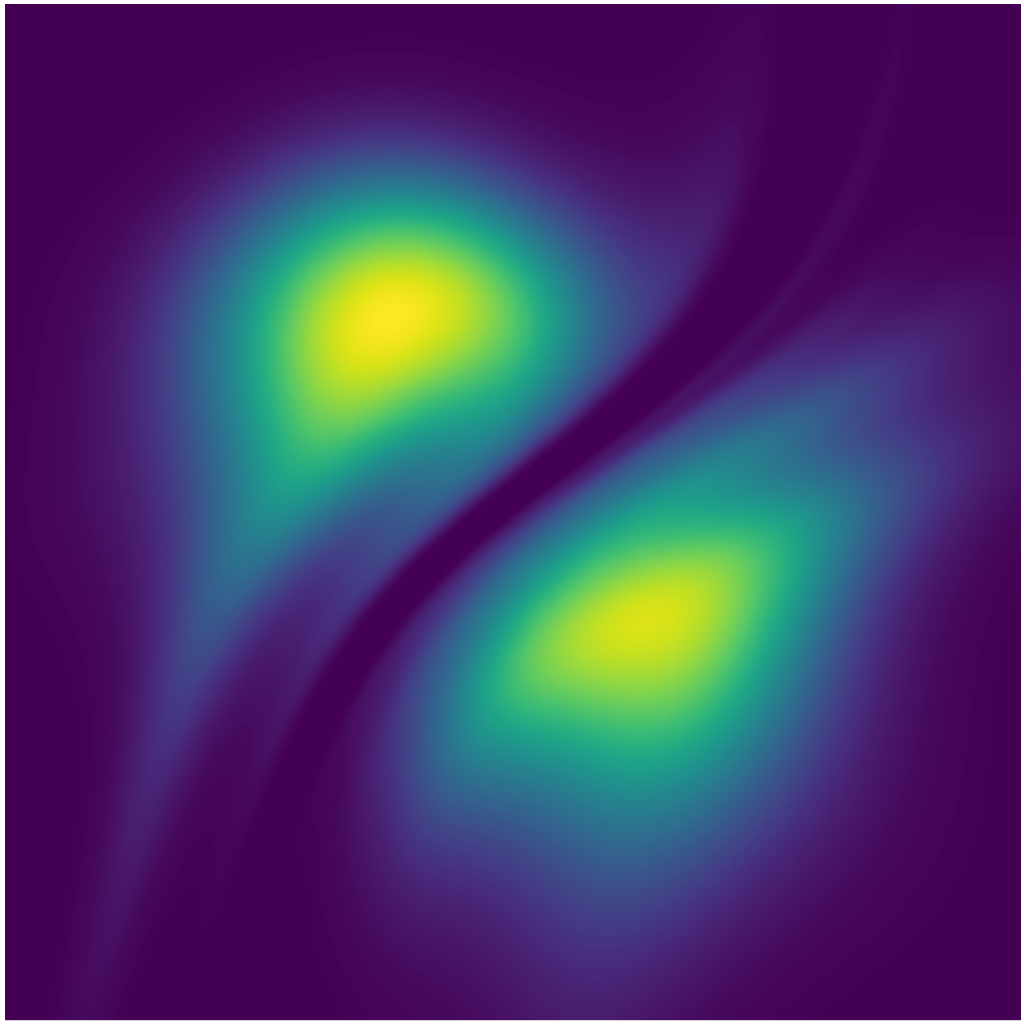}
    & 
    \includegraphics[width=0.3\linewidth]{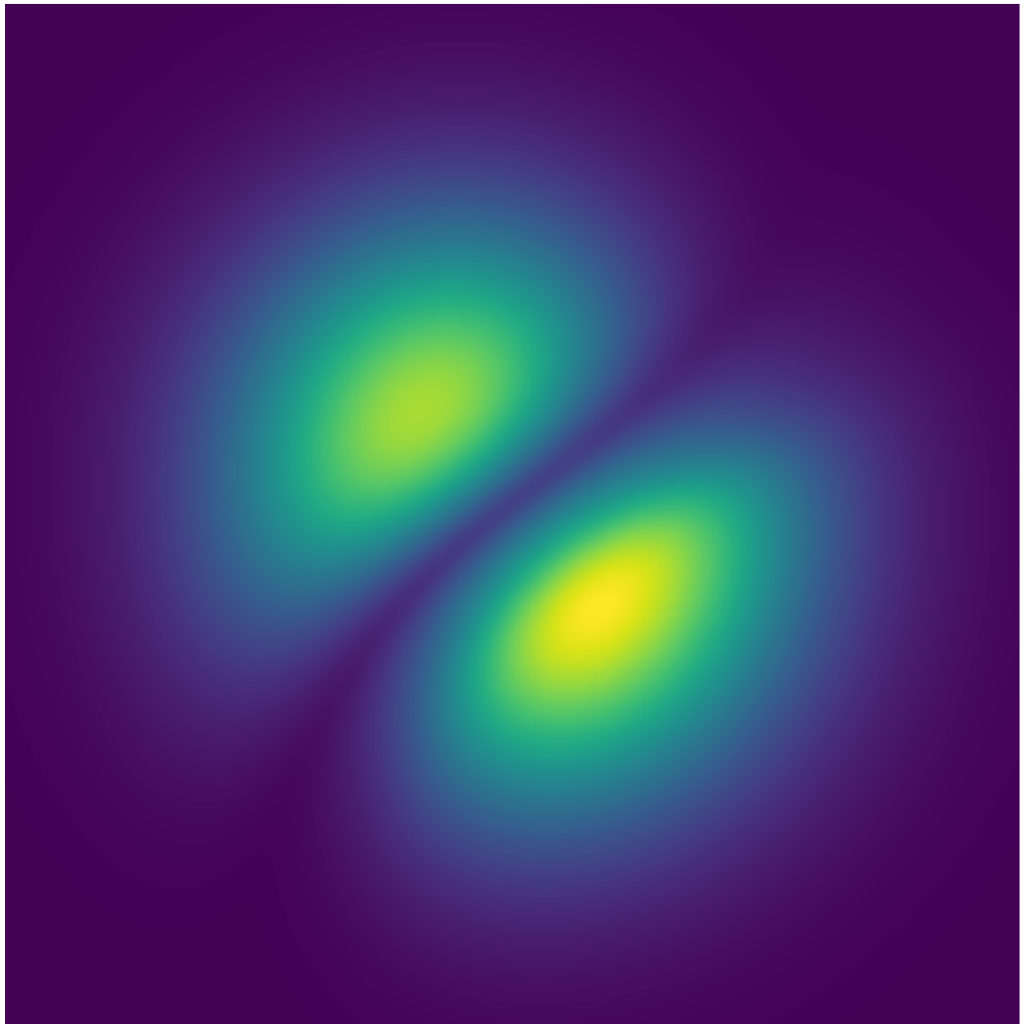}
    &
    \includegraphics[width=0.3\linewidth]{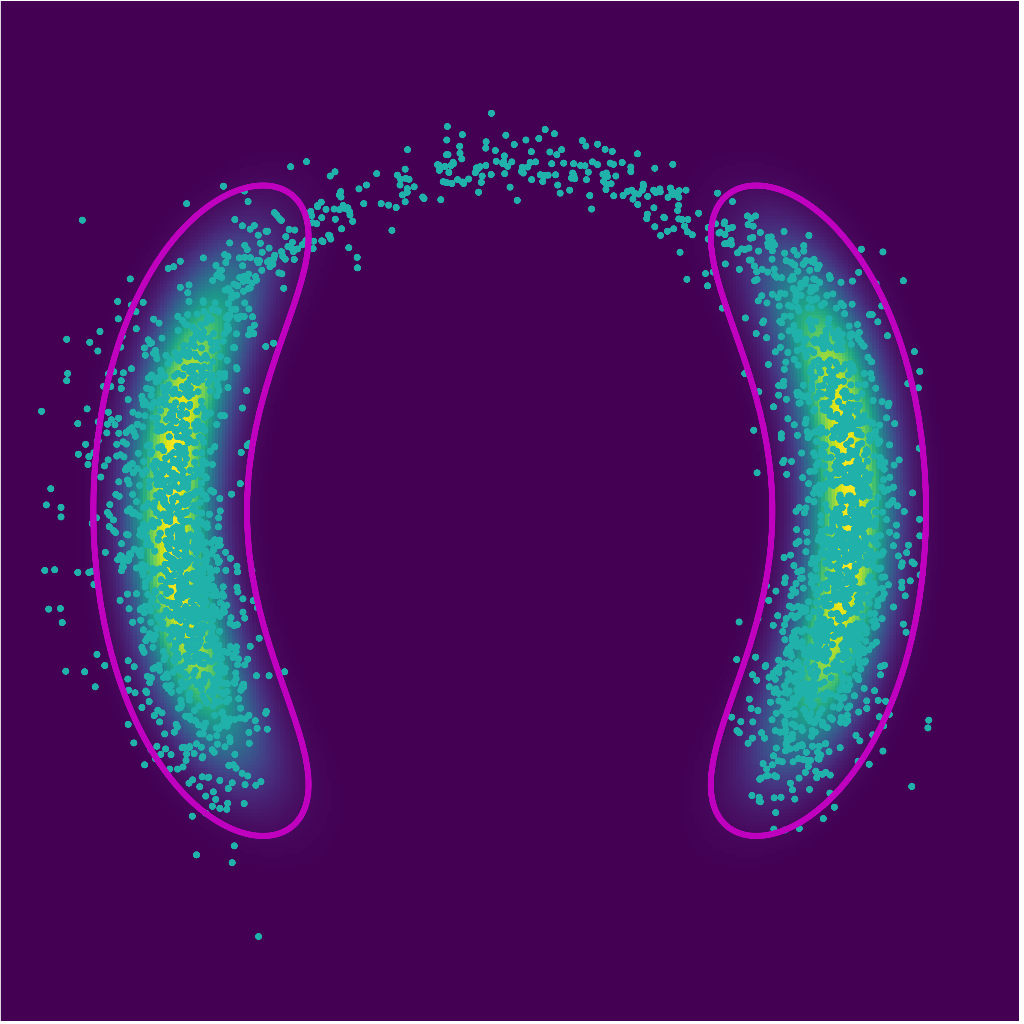}\\
    (a) & (b) & (c) 
    \end{tabular}
    \caption{(a) Target distribution $p_Z$ in the latent space; (b) Learnt latent distribution in the latent space $\hat{q}_Z$; (c) Output of naive sampling, i.e., $\left\{x^{(n)}\right\}_{n=1}^N$ with $x^{(n)} = {f}(z^{(n)})$ and $z \sim q_Z$.}
    \label{fig:jacobian}
\end{figure} 

\par 
Fig. \ref{fig:jacobian} illustrates this behavior on a archetypal bimodal 2D two-moon distribution using a latent Gaussian distribution. The NF is first trained on a large set of samples from the true target distribution. 
%
Fig. \ref{fig:jacobian}(a) shows the likelihood of the inverse transformation of the target distribution back to the latent space. It appears that the NF splits the Gaussian latent space into two sub-regions separated by an area of minimal likelihood. In the target domain, these minima correspond to the low probability area between the two modes of the target distribution, which is somewhat expected.

Fig. \ref{fig:jacobian}(c) shows the result of a naive sampling from the Gaussian model latent distribution to explore the target distribution thanks to the NF change of variable. Latent Gaussian independent samples are transformed by the NF to the target domain. The purple line represents the $97.5\%$ level set. It appears that many samples through this naive sampling procedure are out-of-distribution in the low probability area between the two moons, see the top of the plot. They correspond to samples of the latent distribution around the low-likelihood area highlighted on Fig. \ref{fig:jacobian}(a). Note that this illustrative example and in particular Fig. \ref{fig:jacobian}(b) will be more deeply discussed in the contributive Sections \ref{sec:preliminaries} to \ref{sec:langevin} in the light of the findings reported along the paper.

The observations made above illustrate a behavior that is structural. NFs are diffeomorphisms that preserve the topological structure of the support of the latent distribution. If the information about the structure of the target distribution is ignored, many out-of-distribution samples will be generated. This effect is reinforced by the fact that the NF is trained on a finite data set so that in practice there exist close to empty areas in low probability regions. In other words, since the latent distribution is usually a simple Gaussian unimodal distribution, there is a topological mismatch with the often much more complex target distribution \citep{cornish2020relaxing}, in particular when it is multimodal. 

A first contribution of this paper is a theoretical study of the impact of a topological mismatch between the latent distribution on the Jacobian of the NF transformation. We prove that the norm of the Jacobian of a sequence of differentiable mappings between a unimodal distribution and a distribution with disconnected support diverges to infinity (see Proposition~\ref{prop:jacobian}). This observation suggests that one should consider the information brought by the Jacobian when sampling from the target distribution with a NF.
We also draw a connection between NF optimality and Gaussian isoperimetric theory \citep{ledoux1996isoperimetry}. This connection permits to explain the split of the latent space with minimal perimeter corresponding to Voronoi cells \citep{milman2022gaussian}.

Capitalizing on this theoretical study, the second contribution of this paper is a new dedicated Markov chain Monte Carlo algorithm to sample efficiently according to the distribution targeted by a NF. The proposed sampling method builds on a Langevin dynamics formulated in the target domain and translated into the latent space, which is made possible thanks to the invertibility of the NF. Interestingly the resulting Langevin diffusion is defined on the Riemann manifold whose geometry is driven by the Jacobian of the NF. As a result, the proposed Markov chain Monte Carlo method is shown to avoid low probability regions and to produce significantly less out-of-distribution samples, even when the target distribution is multimodal. It is worth noting that the proposed method does not require a specific training procedure but can be implemented to sample from any pre-trained NF with any architecture.  

The paper is organized as follows. Section~\ref{sec:related_works} reports on related works. Section~\ref{sec:preliminaries} recalls the main useful notions about normalizing flows. Section~\ref{sec:topological_mismatch} studies the theoretical implications of a topological mismatch between the latent distribution and the target distribution. Section~\ref{sec:langevin} introduces the proposed sampling method based on a Langevin dynamics in the latent space. In Section~\ref{sec:experiments}, numerical experiments illustrate the advantages of the proposed approach by reporting performance results both for 2D toy distributions and in high dimensions on the usual Cifar-10, CelebA and LSUN data sets.

\section{Related works}
\label{sec:related_works}

\noindent \textbf{Geometry in neural networks --}  Geometry in neural networks as a tool to understand local generalization was first discussed by \citet{bengio2013representation}. As a key feature, the Jacobian matrix controls how smoothly a function interpolates a surface from some input data. As an extension, \citet{rifai2011higher} showed that the norm of the Jacobian acts as a regularizer of the deterministic autoencoder. Later \citet{arvanitidis2018latent} were the first to establish the link between push forward generative models and surface modeling. In particular, they showed that the latent space could reveal a distorted view of the input space that can be characterized by a stochastic Riemannian metric governed by the local Jacobian.\\

\noindent \textbf{Distribution with disconnected support --}  As highlighted by \citet{cornish2020relaxing}, when using ancestral sampling, the structure of the latent distribution should fit the unknown structure of the target distribution. To tackle this issue, several solutions have been proposed. These strategies include augmenting the space on which the model operates \citep{huang2020augmented}, continuously indexing the flow layers \citep{cornish2020relaxing}, and including stochastic \citep{wu2020stochastic} or surjective layers \citep{nielsen2020survae}. However, these approaches sacrifice the bijectivity of the flow transformation. In most cases, this sacrifice has dramatic impacts: the model is no longer tractable, memory savings during training are no longer possible \citep{gomez2017reversible}, and the model is no longer a perfect encoder-decoder pair. 
Other works have promoted the use of multimodal latent distributions \citep{izmailov2020semi,ardizzone2020training,hagemann2021stabilizing}. Nevertheless, rather than capturing the inherent multimodal nature of the target distribution, their primary motivation is to perform a classification task or to solve inverse problems with flow-based models. \citet{papamakarios2017masked} has shown that choosing a mixture of Gaussians as a latent distribution could lead to an improvement of the fidelity to multimodal distributions. Alternatively, \citet{pires2020variational} has studied the learning of a mixture of generators. Using a mutual information term, they encourage each generator to focus on a different submanifold so that the mixture covers the whole support. More recently, \citet{stimper2022resampling} predicted latent importance weights and proposed a sub-sampling method to avoid the generation of the most irrelevant samples. However, all these methods require to implement elaborated learning strategies which handle several sensitive hyperparameters or impose specific neural architectures. On the contrary, as emphasized earlier, the present approach does not require a specific training strategy, is computationally efficient, and can be implemented to any pre-trained NF.\\

\noindent \textbf{Sampling with normalizing flows --} 
Recently NFs have been used to facilitate the sampling from distributions with non-trivial geometries by transforming them into distributions that are easier to handle. To solve the problem, samplers that combine Monte Carlo methods with NF have been proposed. On the one hand, flows have been used as reparametrization maps that improve the geometry of the target distribution before running local conventional samplers such as Hamiltonian Monte Carlo
(HMC) \citep{hoffman2019neutra, noe2019boltzmann}. On the other hand, the push-forward of the NF base distribution through the map has also been used as an independent proposal in importance sampling \citep{muller2019neural} and Metropolis-Hastings steps \citep{gabrie2022adaptive, samsonov2022local}. In this context, NFs are trained using the reverse Kullback-Leiber divergence so that the push-forward distribution approximates the target distribution. These approaches are particularly appealing when a closed-form expression of the target distribution is available explicitly. In contrast, this paper does not assume an explicit knowledge of the target distribution. The proposed approach aims at improving the sampling from a distribution learnt by a given NF trained beforehand. 

\section{Normalizing flows: preliminaries and problem statement}\label{sec:preliminaries}

\subsection{Learning a change of variables}
%
%
%
%
%

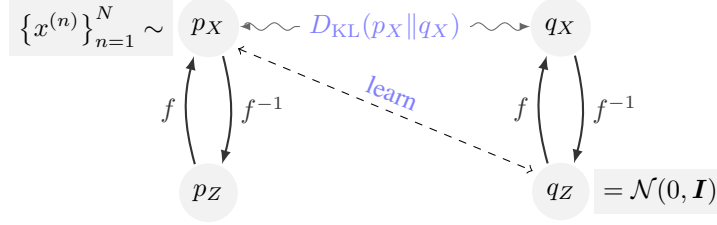
\begin{figure}
\centering
	\begin{tikzpicture}[prob/.style={circle, fill=black!5}, data/.style={rectangle, fill=black!5}]

	\node[prob]        (px)                                  {$p_X$};
	\node[text=blue!50](t1)       [right=0.8cm of px]        {$D_{\mathrm{KL}}(p_X \| q_X)$};
	\node[prob]        (pz)       [below=1.4cm of px]        {$p_Z$};

	\node[prob]        (qx)       [right=0.8cm of t1]        {$q_X$};
	\node[prob]        (qz)       [below=1.4cm of qx]        {$q_Z$};
	\node[data] (NZeroI) [right=0.05cm of qz] {$= \mathcal{N}(0,\boldsymbol{I})$};
    \node[data] (samples) [left=0.05cm of px] {$\left\{x^{(n)}\right\}_{n=1}^{N} \sim $};
	
	\draw[latex-, thick, black!80] ([xshift=0.1cm, yshift=-0.1cm]px.south west) to[bend right=15] node[left] {$f$} ([xshift=0.1cm, yshift=0.1cm]pz.north west);
	\draw[-latex, thick, black!80] ([xshift=-0.1cm, yshift=-0.1cm]px.south east) to[bend left=15] node[right] {$f^{-1}$} ([xshift=-0.1cm, yshift=0.1cm]pz.north east);

	\draw[-latex, thick, black!80] ([xshift=-0.1cm, yshift=-0.1cm]qx.south east) to[bend left=15] node[right] {$f^{-1}$} ([xshift=-0.1cm, yshift=0.1cm]qz.north east);
	\draw[latex-, thick, black!80] ([xshift=0.1cm, yshift=-0.1cm]qx.south west) to[bend right=15] node[left] {$f$} ([xshift=0.1cm, yshift=0.1cm]qz.north west);

	\draw[latex-, black!60, snake it] (px) to (t1);
	\draw[-latex, black!60, snake it] (t1) to (qx);
	\draw[<->, dashed] ([xshift=0.1cm]px.south east) to node[midway, above, sloped, blue!50] {learn} ([xshift=-0.1cm]qz.north west);

\end{tikzpicture}
\caption{NF learns a mapping $f$ from data points $\left\{x^{(n)}\right\}_{n=1}^N$ assumed to be drawn from $p_X$ towards the latent Gaussian measure $q_Z$. The training consists in minimizing the KL divergence between $p_X$ and $q_X=f_{\sharp}q_Z$. Once trained, the learnt map permits to go from $q_Z$ to $q_X$, which is an approximation of the true target distribution $p_X$.}\label{fig:training_NF}
\end{figure}

NFs define a flexible class of deep generative models that seek to learn a change of variable between a reference Gaussian measure $q_Z$ and a target measure $p_{X}$ through an invertible transformation $f: \mathcal{Z} \rightarrow \mathcal{X}$ with $f \in \mathcal{F}$ 
where $\mathcal{F}$ defines the class of NFs. Fig. \ref{fig:training_NF} summarizes the usual training of NF that minimizes a discrepancy measure between the target measure $p_X$ and the push-forwarded measure $q_X$ defined as
\begin{equation} \label{eq:NF_goal_bis}
    q_X = f_{\sharp}q_Z
\end{equation}
where  $ f_{\sharp}$ stands for the associated push-forward operator.
This discrepancy measure is generally chosen as the Kullback-Leibler (KL) divergence $D_{\mathrm{KL}}(p_X \| q_X)$.  Explicitly writing the change of variables 
\begin{equation}\label{eq:change_of_variable}
    q_X(x) = q_Z(f^{-1}(x)) \left|J_{f^{-1}}(x)\right|
\end{equation}
where $J_{f^{-1}}$ is the Jacobian matrix of $f^{-1}$, the training is thus formulated as the minimization problem 
\begin{equation}
\min_{f \in \mathcal{F}} \quad \mathbb{E}_{p_X}[-\log q_Z(f^{-1}(x))+\log |J_{f^{-1}}(x)|]
\label{eq:continious-opt}
\end{equation}
Note that the term $\log p_X(x)$ does not appear in the objective function since this latter does not depend on $f$. In the present work, the class $\mathcal{F}$ of admissible transformations is chosen as the structures composed of coupling layers (\cite{papamakarios2021normalizing, dinh2016density, kingma2018glow}) ensuring the Jacobian matrix of $f$ to be lower triangular with positive diagonal entries. Because of this triangular structure, the Jacobian $J_f$ and the inverse of the map $f^{-1}$ are available explicitly. In particular the Jacobian determinant $\left|J_f(z)\right|$ evaluated at $z\in\mathcal{Z}$ measures the dilation, the change of volume of a small neighborhood around $z$ induced by $f$, i.e., the ratio between the volumes of the corresponding neighborhoods of $x$ and $z$.
 
In practice, the target measure $p_X$ is available only though observed samples $\left\{x^{(1)}, x^{(2)}, \ldots, x^{(N)}\right\}$. Adopting a sample-average approximation, the objective function in \eqref{eq:continious-opt} is replaced by its Monte Carlo estimate. For this fixed set of samples per data batch, the NF training is formulated as
\begin{equation}
\hat{f} \in \min _{f \in \mathcal{F}} \frac{1}{N} \sum_{n=1}^N\left[-\log q_Z(f^{-1}(x^{(n)}))+\log |J_{f^{-1}}(x^{(n)})| \right].
\label{eq:empirical-opt}
\end{equation}
In the following, it will be worth keeping in mind that the obtained solution $\hat{f}$ is only an approximation of the exact transport map due to two reasons. First, the feasible set $\mathcal{F}$ (i.e., the class of admissible NFs) is reduced to the set of continuous, differentiable and bijective functions. The existence of a transformation belonging to this class such that $D_{\mathrm{KL}}(p_X \| q_X)=0$ is not guaranteed. This remark holds even in the case of a parametric form of the target distribution $p_X$, i.e., when minimizing the real (non-empirical) objective function in \eqref{eq:continious-opt}. Second, even when such a transformation exists in $\mathcal{F}$, the solution $\hat{f}$ recovered by \eqref{eq:empirical-opt} coincides with the minimizer of \eqref{eq:continious-opt} in an asymptotic sense only, i.e., when $N\rightarrow \infty$. On top of that, only a proxy of $\hat{f}$ can be computed because of the use of a stochastic optimization procedure (e.g., stochastic gradient descent), which may suffer from issues not discussed anyfurther here.

The main issues inherent to the NF training and identified above would still hold for more refined training procedures  \citep{Coeurdoux_ESANN_2022}, i.e., that would go beyond to the crude minimization problem \eqref{eq:continious-opt}. However, the work reported in this paper does not address the training of the NF. Instead, one will focus on the task which consists in generating samples from the learnt target measure. Thus one will assume that a NF has been already trained to learn a given change of variable. To make the sequel of this paper smoother to read, no distinction will be made between the sought transformation and its estimate, that will be denoted $f$ in what follows. 

\subsection{A Gaussian latent space?}
\label{subsec:Gaussian_space}
As noticed by \citet{marzouk2016sampling}, learning the transformation $f$ by variational inference can be reformulated with respect to (w.r.t.) the corresponding inverse map $f^{-1}$. Since the KL divergence is invariant to changes of variables, minimizing $D_{\mathrm{KL}}(p_X\|q_X)$ is equivalent to minimizing $D_{\mathrm{KL}}(p_Z \| q_Z)$ with $p_Z=f_{\sharp}^{-1} p_X$. The training procedure is thus formulated in the latent space instead of the target space. In other words, the NF aims at fitting the target measure $p_Z$ expressed in the latent space to the latent Gaussian measure $q_Z$. However, due to inescapable shortcomings similar to those highlighted above, the target measure $p_Z$ in the latent space is only an approximation of the latent Gaussian measure $q_Z$. This mismatch can be easily observed in Fig. \ref{fig:jacobian}(a) where the depicted actual measure $p_Z$ is clearly not Gaussian. This issue may be particularly critical when there is a topological mismatch between the respective supports of the target and latent distributions. This will be discussed in more details in Section \ref{sec:topological_mismatch}. 

\subsection{Beyond conventional NF sampling}
\label{subsec:NF_sampling}
Once the NF has been trained, the standard method to sample from the learnt target distribution is straightforward. It consists in drawing a sample $z_k$ from the latent Gaussian distribution $q_Z$ and then applying the learnt transformation $f$ to obtain a sample $x^{(n)}=f\left(z^{(n)}\right)$. This method will be referred to as ``naive sampling'' in the sequel of this paper.

Unfortunately, as discussed in Section \ref{subsec:Gaussian_space} (see also Fig. \ref{fig:jacobian}), the latent distribution $q_Z$ is expected to be different from the actual target distribution $p_Z$ expressed in the latent space. As suggested in the next section, this mismatch will be even more critical when it results from topological differences between the latent and target spaces. As a consequence the naive NF sampling is doomed to be suboptimal and to produce out-of-distribution samples, as illustrated in Fig. \ref{fig:jacobian}(c). In contrast, the approach proposed in Section \ref{sec:langevin} aims at devising an alternative sampling strategy that explicitly overcomes these shortcomings.

\section{Implications of a topological mismatch}
\label{sec:topological_mismatch}

\subsection{Topology preservation}\label{subsec:topology_preservation}

The push-forward operator $f_\sharp$ learnt by an NF transports the mass allocated by $q_Z$ in $\mathcal{Z}$ to $\mathcal{X}$, thereby defining $q_X$ by specifying where each elementary mass is transported. This imposes a global constraint on the operator $f$ if the model distribution $q_X$ is expected to match a given target measure $p_X$ perfectly. Let $\mathrm{supp}(q_Z)=\{z \in \mathcal{Z}: q_Z(z)>0\}$ denote the support of $q_Z$. Then the push-forward operator $f_\sharp$ can yield $q_X=p_X$ only if
\begin{equation}\label{eq:support_constraint}
\operatorname{supp}(p_X)=\overline{f\left(\operatorname{supp}(q_Z)\right)}
\end{equation}
where $\overline{B}$ is the closure of set $B$.
The constraint \eqref{eq:support_constraint} is especially onerous for NFs because of their bijectivity. The operators $f$ and $f^{-1}$ are continuous, and $f$ is a homeomorphism. Consequently, for these models, $q_Z$ and $p_X$ are isomorphic, i.e., homeomorphic as topological spaces \citep[Def. 3.3.10]{runde2005taste}. This means that $\operatorname{supp}(q_Z)$ and $\operatorname{supp}(p_X)$ must share exactly the same topological properties, in particular the number of connected components. 
This constraint may be unlikely satisfied when learning complex real-world distributions, leading to an insurmountable topological mismatch. In such cases, this finding has serious consequences on the operator $f$ learnt and implemented by a NF. Indeed, the following  proposition states that if the respective supports of the latent distribution $q_Z$ and the target distribution $p_X$ are not homeomorphic, then the norm of the Jacobian $|J_f|$ of $f$ may become  arbitrary large. Here $\stackrel{\mathcal{D}}{\rightarrow}$ denotes weak convergence.

\begin{prop}\label{prop:jacobian}
Let $q_Z$ and $p_X$ denote distributions defined on 
$\mathbb{R}^d$. Assume that $\text{supp}(q_Z) \neq \text{supp}(p_X)$. 
For any sequence of measurable, differentiable Lipschitz functions $f_t : \mathbb{R}^{d_Z} \rightarrow \mathbb{R}^{d_X}$, if $f_{t\sharp} q_Z \xrightarrow{\mathcal{D}} p_X$ when ${t\rightarrow +\infty}$, then
$$
 \lim _{t \rightarrow \infty} \sup _{z \in \mathcal{Z}}( \left \| J_{f_t}(z) \right \|) = +\infty.
$$
\end{prop}

The proof is reported in Appendix \ref{sec:proof}.\\

It is worth noting that training a generative model is generally conducted by minimizing a  statistical divergence. For most used divergence measures, (e.g., KL and Jensen-Shannon divergences,  Wasserstein distance), this minimization implies a weak convergence of the approximated distribution $q_X$ towards the target distribution $p_X$ \citep{arjovsky2017wasserstein}. As a consequence, Proposition \ref{prop:jacobian} states that, when training a NF to approximate $p_X$ by $q_X = \lim_{t \rightarrow \infty} f_{t\sharp} q_Z$, the supremum of the Jacobian of the learnt mapping may become arbitrarily large in some regions. This result is in line with the experimental findings early discussed and visually illustrated by Fig. \ref{fig:jacobian}. Indeed, Fig. \ref{fig:jacobian}(b) depicts the heatmap of the log-likelihood  
\begin{equation}
    \log q_X(f(z)) = \log q_Z(z) - \log \left | J_{f}(z) \right | 
    \label{eq:hatpz}
\end{equation}
given by \eqref{eq:change_of_variable}  after training an NF. 
The impact of the term governed by the determinant of the Jacobian is clear. It highlights a boundary separating two distinct areas, each associated with a mode in the target distribution $p_X$. This result still holds when $q_Z$ and $q_X$ are defined on $\mathbb{R}^{d_Z}$ and $\mathbb{R}^{d_X}$, respectively, with $d_Z \neq d_X$. This shortcoming is thus also unavoidable when learning injective flow models \citep{kumar2017semi} and other push-forward models such as GANs \citep{goodfellow2020generative}.


In practice, models are trained on a data set of finite size. In other words, the underlying target measure $p_X$ is available only through the empirical measure $\frac{1}{N}\sum_{n=1}^{N} \delta_{x^{(n)}}$. During the training defined by \eqref{eq:empirical-opt}, areas of low probability possibly characterizing a multi-modal target measure are likely interpreted as areas of null probability observed in the empirical measure. This directly results in the topological mismatch discussed above. Thus, even when targeting a distribution $p_X$ defined over a connected support with regions of infinitesimal support, the learnt mapping is expected to be characterized by a Jacobian with exploding norm in these regions, see Fig \ref{fig:jacobian}.  The following section suggests that these regions correspond to the frontiers between cells defining a partition of the latent space.


\subsection{Partition of the latent space}

A deterministic generative model can be interpreted as a surface model or a Gauss map, if the generator $f$ is sufficiently smooth \citep{arvanitidis2018latent}. When targeting a  multi-modal distribution, the learnt model implicitly partitions the Gaussian latent space into a set of disjoint subsets, each associated with a given mode. The Gaussian multi-bubble conjecture was formulated when looking for a way to partition the Gaussian space with the least-weighted perimeter. This conjecture was proven recently by \citet{milman2022gaussian}. The result states that partitioning a Gaussian space of $\mathbb{R}^d$ into $m$ clusters of equal measures ($2 \leqslant m \leqslant d+1$) consists in recovering ``simplicial clusters" defined as Voronoi cells of $m$ equidistant points in $\mathbb{R}^d$. According to convex geometry principles, the boundary is the union of $m$ convex cones, each of them contained in a distinct hyperplane that goes through the origin of $\mathbb{R}^d$. Recently, \citet{issenhuth2022optimal} leveraged on this finding to assess the optimality of the precision of GANs. They show that the precision of the generator vanishes when the number of components of the target distribution tends towards infinity. 

Again, in practice, NFs models are trained on a data set with a finite number of samples. This results in a partitioning of the Gaussian latent space into cells separated by frontiers of arbitrarily large widths. Figure  \ref{fig:jacobian}(b) allows a connection to be drawn between the statement by  \citet{milman2022gaussian} and the Proposition \ref{prop:jacobian}. Indeed, in this figure, the frontiers defining this partitioning are clearly identified as the areas with exploding Jacobian norm. As a consequence, exploring naively the Gaussian latent space to sample from the target distribution seems to be inappropriate. Because of these wide frontiers, large areas of the latent space are expected to be associated with potentially numerous out-of-distribution samples.

\section{NF sampling in the latent space}
\label{sec:langevin}
\subsection{Local exploration of the latent space}

As explained in Section \ref{subsec:NF_sampling}, naive NF sampling boils down to drawing a Gaussian variable before transformation by the learnt mapping $f$. This strategy is expected to produce out-of-distribution samples, due to the topological mismatch between $q_X$ and $p_X$ discussed in Section \ref{sec:topological_mismatch}. The proposed alternative elaborates directly on the learnt target distribution $q_X$.

The starting point of our rational consists in expressing a Langevin diffusion in the target space. This Markov chain Monte Carlo (MCMC) algorithm would target the distribution $q_X$ using only the derivative of its likelihood $\nabla_x\log q_X({x})$. After initializing the chain by drawing from an arbitrary distribution ${x}_0 \sim \pi_0(x)$, the updating rule writes
\begin{equation}
x_{k+1} \leftarrow x_k+\frac{\epsilon^2}{2} \nabla_{x} \log q_X(x_k)+ \epsilon  \xi
\label{eq:langevin}
\end{equation}
where $\xi \sim \mathcal{N}(0, I)$. When $\epsilon \rightarrow 0$ and the number of samples $K \rightarrow \infty$, the distribution of the samples generated by the iterative procedure  \eqref{eq:langevin} converges to $q_X$ under some regularity conditions. In practice, the error is negligible when $\epsilon$ is sufficiently small and $K$ is sufficiently large.  This algorithm referred to as the unadjusted Langevin Algorithm (ULA) always accepts the generated sample proposed by \eqref{eq:langevin}, neglecting the errors induced by the discretization scheme of the continuous diffusion. To correct this bias, Metropolis-adjusted Langevin Algorithm (MALA) applies a Metropolis-Hastings step to accept or reject a sample proposed by ULA \citep{grenander1994representations}.

Again, sampling according to $q_X$ thanks to the diffusion \eqref{eq:langevin} is likely to be inefficient due to the expected complexity of the target distribution possibly defined over a  subspace of $\mathbb{R}^d$. In particular, this strategy suffers from the  lack of prior knowledge about the location of the mass. 
Conversely,  the proposed approach explores the latent space by leveraging on the closed-form change of variable \eqref{eq:change_of_variable} operated by the trained NF. 
After technical derivations reported in Appendix \ref{subsec:appendix_derivation_proposal}, the counterpart of the diffusion \eqref{eq:langevin} expressed in the latent space writes
\begin{equation}    
z^{\prime} = z_k + \frac{\epsilon^2}{2} G^{-1}(z_k) \nabla_{z} \log \tilde{q}_Z(z_k) +   \epsilon \sqrt{G^{-1}(z_k)}  \xi
\label{eq:latentlangevin}
\end{equation}
where
\begin{equation}\label{eq:q_tilde}
    \tilde{q}_Z(z) = q_Z(z) \left|J_{f}(z)\right|^{-1}
\end{equation}
and
\begin{equation}
   G^{-1}(z)  = \left[J_f^{-1}(z)\right]^2. \label{eq:G}
\end{equation}
%
Note that the distribution $\tilde{q}_Z$ in \eqref{eq:q_tilde} originates from the change of variable that defines $q_X$ in \eqref{eq:change_of_variable} and has been already implicitly introduced by \eqref{eq:hatpz} in Section \ref{sec:topological_mismatch}. Interestingly, the matrix $G(\cdot)$ is a definite positive matrix (see Appendix \ref{app:sec:jacobian}).
Thus the diffusion \eqref{eq:latentlangevin} characterizes a Riemannian manifold Langevin dynamics where $G(\cdot)$ is the Riemannian metric associated with the latent space  \citep{girolami2011riemann, xifara2014langevin}. More precisely, it defines the conventional proposal move of the Riemannian manifold ajusted Langevin algorithm (RMMALA) which targets the distribution $\tilde{q}_Z$ defined by \eqref{eq:q_tilde}.
This distribution is explicitly defined through the Jacobian $J_{f}(\cdot)$ of the transformation whose behavior has been discussed in depth in Section \ref{sec:topological_mismatch}. It can be interpreted as the Gaussian latent distribution $q_Z$ tempered by the (determinant of the) Jacobian of the transformation. It has also been evidenced by depicting the heatmap of \eqref{eq:hatpz} in Fig. \ref{fig:jacobian}(b), which shows that it appears as a better approximation of $p_Z$ than $q_Z$. Since it governs the drift of the diffusion through the gradient of its logarithm, the diffusion is expected to escape from the areas where the determinant of the Jacobian explodes, see Section \ref{sec:topological_mismatch}.

The proposal kernel $g(z^{\prime}|z)$ associated with the diffusion \eqref{eq:latentlangevin} is a Gaussian distribution whose probability density function (pdf) can be conveniently rewritten as (see Propriety \ref{prop:kernel} in Appendix \ref{subsec:appendix_derivation_proposal})
\begin{equation}
    g\left(z^{\prime} \mid z_k\right)  \propto |J_{f(z_k)}|   \exp \left[-\frac{1}{2 \epsilon^{2}} \left \| J_{f}(z_k)(z^{\prime}-z_k) + \frac{\epsilon^2}{2} \ts_Z(z_k) \right \|^2\right].
\label{proposal}
\end{equation}
where $\ts_Z(\cdot)$ denotes the so-called latent score
\begin{equation}
    \ts_Z(z) = J^{-1}(z) \nabla_{z} \log \tilde{q}_Z(z).
    \label{eq:latent_score}
\end{equation}
The sample proposed according to \eqref{eq:latentlangevin} is then accepted with probability
\begin{equation}
    \alpha_{\mathrm{RMMALA}}(z_k,z^{\prime}) =\min \left(1, \frac{\tilde{q}_Z\left(z^{\prime}\right) g\left(z_k \mid z^{\prime}\right)}{\tilde{q}_Z\left(z_k\right) g\left(z^{\prime} \mid z_k\right)}\right).
    \label{eq:MALA-alp}
\end{equation}
It is worth noting that the formulation \eqref{proposal} of the proposal kernel leads to a significantly faster implementation than its canonical formulation. Indeed, it does not require to compute the metric $G^{-1}(\cdot)$ defined by \eqref{eq:G}, which depends on the inverse of the Jacobian matrix twice. Moreover, the evaluation of the latent score \eqref{eq:latent_score} can be achieved in an efficient manner, bypassing the need for evaluating the inverse of the Jacobian matrix, as elaborated in Appendix \ref{app:subsubsec:fast_comput}. Finally, only the Jacobian associated with the forward transformation is required to compute \eqref{proposal}. This approach enables a streamlined calculation of the acceptance ratio \eqref{eq:MALA-alp}, ensuring an overall computational efficiency.

Besides, the proposal scheme \eqref{eq:latentlangevin} requires to generate high dimensional Gaussian variables with covariance matrix $\epsilon^2 G^{-1}(\cdot) $ \citep{Vono_SIREV_2022}. To lighten the corresponding computational burden, we take advantage of a $1$st order expansion of $f^{-1}$
to approximate \eqref{eq:latentlangevin} by the diffusion (see Appendix \ref{app:subsubsec:approximation_expansion})
\begin{equation}
z^{\prime} = f^{-1}\left(f(z_k) +\epsilon \xi \right) + \frac{\epsilon^2}{2} J_f^{-1}(z_k) \ts_Z(z).
\label{eq:diffusion_eff}
\end{equation}
According to \eqref{eq:diffusion_eff}, this alternative proposal scheme  requires to generate high dimensional Gaussian variables with a covariance matrix which is now identity, i.e., most cheaper. Moreover, it is worth noting that \emph{i)}  the latent score $\ts_Z(\cdot)$ can be evaluated efficiently (see above) and \emph{ii)} using $J_{f}^{-1}(z)=J_{f^{-1}}(f(z))$ (see Property \ref{prop:jaco_trick} in Appendix \ref{app:subsec:preliminaries}), sampling $z^{\prime}$ according to \eqref{eq:diffusion_eff} only requires to evaluate the Jacobian associated with the backward transformation.
Proofs and implementation details are reported in Appendix \ref{sec:lattent_diff}. The algorithmic procedure to sample according to this kernel denoted $\mathcal{K}_{\mathrm{RMMALA}}(\cdot)$ is summarized in Algo. \ref{al:MALA}.

\begin{algorithm}[h]
    \caption{Sampling kernel $\mathcal{K}_{\mathrm{RMMALA}}(\cdot)$. }
    \label{al:MALA}

\KwIn{trained NF $f(\cdot)$, time step $\epsilon$, current state $z_k$ of the sampler.}
    \tcc{ Draw the candidate} 
    
    Draw $\xi \sim \mathcal{N}(0,1)$
    
    Set $z^{\prime} = f^{-1}\left(f(z_k) +  \epsilon \cdot \xi \right)+\frac{\epsilon^2}{2} J_f^{-1}(z_k) \ts_{Z}(z_k)$
    
    \tcc{ Accept/reject procedure} 
    
    Draw $u \sim \mathcal{U}(0,1)$ 
    
    \uIf{$u < \alpha_{\mathrm{RMMALA}}(z_k,z^{\prime})$}{
        Set $z_{k+1} = z^{\prime}$ 
      }
    \uElse{
        Set $z_{k+1} = z_{k}$ 
      }
\KwOut{New state $z_{k+1}=\mathcal{K}_{\mathrm{RMMALA}}(z_k)$ of the sampler.}          
\end{algorithm}

\subsection{Independent Metropolis-Hastings sampling}
Handling distributions that exhibit several modes or defined on a complex multi-component topology is another major issue raised by the problem addressed here. In practice, conventional sampling schemes such as those based on Langevin dynamics fail to explore the full distribution when modes are isolated since they may get stuck around one of these modes. Thus, the samples proposed according to \eqref{eq:latentlangevin} in areas with high values of $\| J_f(\cdot) \|$ are expected to be rejected. These areas have been identified in Section \ref{sec:topological_mismatch} as the low probability regions between modes when targeting a multimodal distribution.  To alleviate this problem, one strategy consists in resorting to 
another kernel to propose moves from one high probability region to another, without requiring to cross the low probability regions. Following this strategy, this paper proposes to combine the diffusion \eqref{eq:latentlangevin} with an independent Metropolis-Hastings (I-MH) with the distribution $q_Z$ as a proposal. The corresponding acceptance ratio writes
\begin{equation}
\begin{split}
\alpha_{\text{I-MH}}(z_k,z^{\prime}) &= \min \left(1, \frac{\tilde{q}_Z\left(z^{\prime}\right) q_Z(z_{k})}{\tilde{q}_Z\left(z_{k}\right) q_Z(z^{\prime})}\right) \\    
&= \min \left(1, \frac{|J_{f}(z_k)|}{|J_{f}(z^{\prime})|}\right).
    \label{eq:MH-alp}
\end{split}
\end{equation}
It is worth noting that this probability of accepting the proposed move only depends on the ratio between the  Jacobians evaluated at the current and the candidate states. In particular, candidates located in regions of the latent space characterized by exploding Jacobians in case of a topological mismatch (see Section \ref{sec:topological_mismatch}) are expected to be rejected with high probability. Conversely, this kernel will favor moves towards other high probability regions not necessarily connected to the regions of the current state. The algorithmic procedure is sketched in Algo. \ref{al:I-MH}.

\begin{algorithm}[h]
    \caption{Sampling kernel $\mathcal{K}_{\mathrm{I-MH}}(\cdot)$. }
    \label{al:I-MH}

\KwIn{trained NF $f(\cdot)$, current state $z_k$ of the sampler.}
        \tcc{Draw candidate}
        
        \text{Draw} $z^\prime \sim \mathcal{N}(0,1)$ 

        \tcc{ Accept/reject procedure} 
        
        \text{Draw} $u \sim \mathcal{U}(0,1)$ 
        
        \uIf{$u < \alpha_{\mathrm{I-MH}}(z_k, z^{\prime} )$}{
                Set $z_{k+1} = z^{\prime}$ 
        }
        \uElse{
                Set $z_{k+1} = z_{k}$ 
            }
\KwOut{New state $z_{k+1}=\mathcal{K}_{\mathrm{I-MH}}(z_k)$ of the sampler.}          
\end{algorithm}

Finally, the overall proposed sampler, referred to as NF-SAILS for NF SAmpling In the Latent Space and summarized in Algo. \ref{al:sampler}, combines the transition kernels $\mathcal{K}_{\mathrm{RMMALA}}$ and $\mathcal{K}_{\mathrm{I-MH}}$, which  permits to efficiently explore the latent space both locally and globally. At each iteration $k$ of the sampler, the RMMALA kernel $\mathcal{K}_{\mathrm{RMMALA}}$ associated with the acceptance ratio \eqref{eq:MALA-alp} is selected with probability $p$ and  the I-MH kernel $\mathcal{K}_{\mathrm{I-MH}}$ associated with acceptance ratio \eqref{eq:MH-alp} is selected with the probability $1-p$. Again, one would like to emphasize that the  proposed strategy does not depend on the NF architecture and can be adopted to sample from any pretrained NF model.

\begin{algorithm}[h]
    \caption{NF-SAILS: NF SAmpling In the Latent Space.}
    \label{al:sampler}
\KwIn{trained NF $f(\cdot)$, time step $\epsilon$, probability $p$}
    \tcc{ Initialization} 

    Draw $z_0\sim \pi_0(z)$

    \For{$k = 0$ \KwTo $K$}{
    \tcc{ Choose the kernel} 
    
    Draw $u \sim \mathcal{U}(0,1)$

    \uIf{$u < p$}{
        \tcc{LOCAL EXPLORATION (see Algo. \ref{al:MALA})}        
        
        $z_{k+1} = \mathcal{K}_{\mathrm{RMMALA}}(z_k)$
        }
    \uElse{
        \tcc{GLOBAL EXPLORATION (see Algo. \ref{al:I-MH})}
        
        $z_{k+1} = \mathcal{K}_{\mathrm{I-MH}}(z_k)$
        }
    }
\KwOut{Collection of samples $\left\{ z_{k} \right\}_{k=1}^K$.}   

\end{algorithm}


\section{Experiments}
\label{sec:experiments}

This section reports performance results to illustrate the efficiency of NF-SAILS thanks to experiments based on several models and synthetic data sets. It is compared to  state-of-the-art generative models known for their abilities to handle multimodal distributions. These results will show that the proposed sampling strategy achieves good performance, without requiring to adapt the NF training procedure or resorting to non-Gaussian latent distributions. We will also confirm the relevance of the method when working on popular image data sets, namely Cifar-10 \citep{krizhevsky2010cifar}, CelebA \citep{liu2015faceattributes} and LSUN \cite{yu2015lsun}.

To illustrate the versatility of proposed approach w.r.t. the NF architecture, two types of coupling layers are used to build the trained NFs. For the experiments conducted on the synthetic data sets, the NF architecture is RealNVP \citep{dinh2016density}. Conversely, a Glow model is used for experiments conducted on the image data sets \citep{kingma2018glow}. However, it is worth noting that the proposed method can apply on top of any generative model fitting multimodal distributions. Additional details regarding the training procedure are reported in Appendix \ref{app:subsec:training}. 

\subsection{Figures-of-merit}

To evaluate the performance of the NFs, several figures-of-merit have been considered. When addressing bi- dimensional problems, we perform a Kolmogorov-Smirnov test to assess the quality of the generated samples w.r.t. the underlying true target distribution \citep{justel1997multivariate}. The goodness-of-fit is also monitored by evaluating the mean log-likelihood of the generated samples and the entropy estimator between samples, which approximates the Kullback-Leibler divergence between empirical samples \citep{kraskov2004estimating}. 

For applications to higher dimensional problems, such as image generation, the performances of the compared algorithms are evaluated using the Fr\'echet inception distance (FID) \citep{heusel2017gans} using a classifier pre-trained specifically on each data set. Besides, for completeness, we report the bits per dimension (bpd) \citep{papamakarios2017masked}, i.e., the log-likelihoods in the logit space, since this is the objective optimized by the trained models.

\subsection{Results obtained on synthetic data set}\label{subsec:results_synth}

As a first illustration of the performance of NF-SAILS, we consider to learn a mixture of $k$ bidimensional Gaussian distributions, with $k\in\left\{2, 3, 4, 6, 9\right\}$. The NF model $f(\cdot)$ is a RealNVP \citep{dinh2016density} composed of $M=4$ flows, each composed of two three-layer neural networks ($d \rightarrow 16 \rightarrow 16 \rightarrow d$) using hyperbolic tangent activation function. We use the Adam optimizer with learning rate $10^{-4}$ and a batch size of $500$ samples. 

\begin{table}
\centering
\begin{tabular}{ |c|c|c|c|c|c| } 
 \cline{3-5}%
  \multicolumn{2}{c|}{}& $\uparrow\log p_X$ & $\downarrow$KL & $\downarrow$KS  \\ 
 \hline
 \multirow{4}{*}{\rotatebox{90}{$k=2$}}&
 Naive sampling & $-4.13$ & 0.263 & 0.177  \\ 
 & NF-SAILS & $\mathbf{-1.41}$ & ${\bf 0.057}$ & ${\bf 0.047}$  \\  
 & WGAN-GP & {N/A} &  0.308 & 0.287  \\ 
 & DDPM & $-2.98$ & 0.121 & 0.066   \\
 \hline
 \multirow{4}{*}{\rotatebox{90}{$k=3$}}
 & Naive sampling & $-3.52$ & 0.914 & 0.248  \\ 
 & NF-SAILS & $\mathbf{-1.83}$ & ${\bf 0.056}$ & $\mathbf{0.034}$  \\ 
 & WGAN-GP & {N/A} & 0.973 & 0.237  \\ 
 & DDPM & $-2.97$ & 0.364 & 0.124   \\ 
 \hline
 \multirow{4}{*}{\rotatebox{90}{$k=4$}}&
 Naive sampling & $-3.08$ & 0.967 & 0.295  \\
 & NF-SAILS & $\mathbf{-1.07}$ & $\mathbf{0.044}$ & ${\bf 0.041}$  \\  
 & WGAN-GP & {N/A} & 1.012 & 0.317  \\ 
 & DDPM & $-1.81$ & 0.427 & 0.127 \\ 
 \hline
 \multirow{4}{*}{\rotatebox{90}{$k=6$}}&
 Naive sampling & $-2.06$ & 1.219 & 0.205  \\ 
 & NF-SAILS & $\mathbf{-1.09}$ & $\mathbf{0.039}$ & ${0.309}$  \\ 
 & WGAN-GP & {N/A} & 1.392 & 0.212  \\ 
 & DDPM & $-1.99$ & 1.004 & {\bf 0.179}   \\ 
 \hline
\multirow{4}{*}{\rotatebox{90}{$k=9$}}&
 Naive sampling & $-2.297$ & 1.764 & 0.215  \\ 
 & NF-SAILS & $\mathbf{-0.801}$ & ${\bf 0.151}$ & $\mathbf{0.052}$  \\ 
 & WGAN-GP & {N/A} & 1.939 & 0.340  \\ 
 & DDPM & $-1.258$ & 0.906 & 0.205   \\  
 \hline
\end{tabular}
\caption{Goodness-of-fit of the generated samples w.r.t. the number $k$ of Gaussians. Reported
scores result from the average over 5 Monte Carlo runs.}
\label{tab:Gaussians}
\end{table}

Table \ref{tab:Gaussians} reports the considered metrics when  comparing the  proposed NF-SAILS sampling method  to a naive sampling (see Section \ref{subsec:NF_sampling}) or to state-of-the-art sampling techniques from the literature, namely Wasserstein GAN with gradient penalty (WGAN-GP) \citep{gulrajani2017improved} and denoising diffusion probabilistic models (DDPM) \citep{ho2020denoising}. 
These results show that NF-SAILS consistently competes favorably against the compared methods, in particular as the degree of multimodality of the distribution increases. Note that WGAN-GP exploits a GAN architecture. Thus, contrary to the proposed NF-based sampling method, it is unable to provide an explicit evaluation of the likelihood, which explains the N/A values in the table.

\begin{figure}
    \centering
    \medskip
    
    \begin{subfigure}[t]{0.24\textwidth}
        \includegraphics[width=\textwidth]{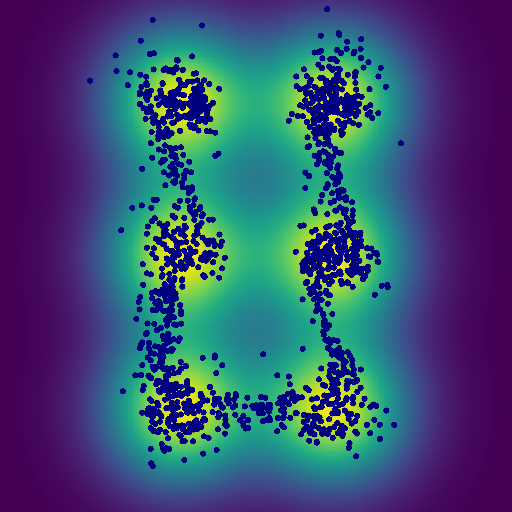}
        \caption{Naive sampling}
        \label{fig:ancestral}
    \end{subfigure}
    \begin{subfigure}[t]{0.24\textwidth}
         \includegraphics[width=\textwidth]{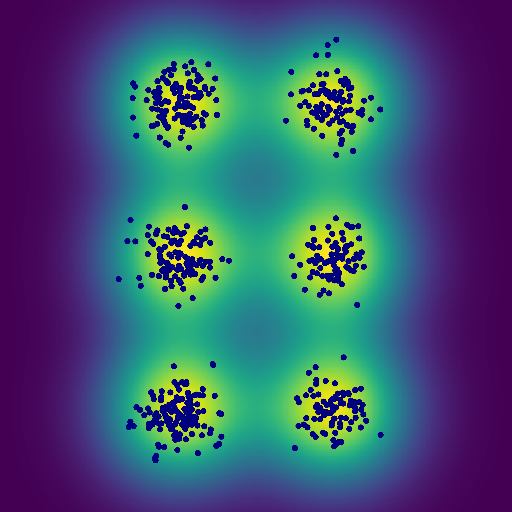}
         \caption{NF-SAILS}
         \label{fig:MALA}
     \end{subfigure}
     \begin{subfigure}[t]{0.24\textwidth}
        \includegraphics[width=\textwidth]{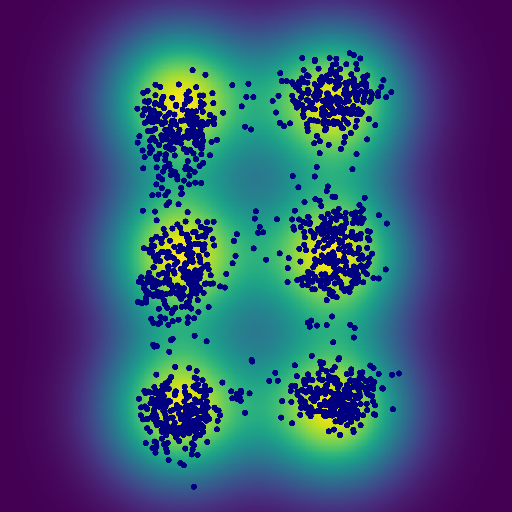}
        \caption{WGAN-GP}
    \end{subfigure}     
    \begin{subfigure}[t]{0.24\textwidth}
        \includegraphics[width=\textwidth]{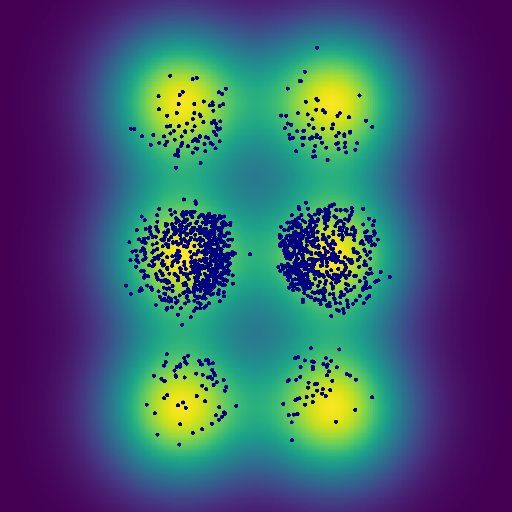}
        \caption{DDPM}
    \end{subfigure}
    \caption{Mixture of $k=6$ Gaussian distributions (green), and 1000 generated samples (blue). The proposed NF-SAILS method in Fig. \ref{fig:MALA}  does not generate samples in-between modes.}
    \label{fig:Gaussians}
\end{figure}

Figure \ref{fig:Gaussians} illustrates this result for $k=6$ and shows that our method considerably reduces the number of out-of-distribution generated samples. Additional results are reported in Appendix \ref{app:sec:complementary}.

\subsection{Results obtained on real image data sets}

Moreover, we further study the performance of NF-SAILS on three different real image data sets, namely Cifar-10 \citep{krizhevsky2010cifar}, CelebA \citep{liu2015faceattributes} and LSUN \citep{yu2015lsun}. Following the same protocol as implemented by \cite{kingma2018glow}, we use a Glow architecture where each neural network are composed of three convolutional layers. The two hidden layers have ReLU activation functions and 512 channels. The first and last convolutions are $3 \times 3$, while the center convolution is $1 \times 1$, since its input and output have a large number of channels, in contrast with the first and last convolutions. Details regarding the implementation are reported in Appendix \ref{app:sec:implementation_details}.

\begin{figure*}
    \begin{minipage}[t]{0.31\columnwidth}
         \centering
        \begin{center}
        \resizebox{\columnwidth}{!}{%
        \begin{tabular}{ |c|c|c| } 
         \cline{2-3}
         \multicolumn{1}{c|}{}   & $\downarrow$ bpd & $\downarrow$ FID \\ 
         \hline
         Naive sampling & 3.35 & 44.6 \\ 
         NF-SAILS & \textbf{3.08} & 43.11 \\ 
         WGAN-GP & N/A & \textbf{18.8} \\
         \hline
        \end{tabular}
        }
        \end{center}
        \includegraphics[width=\textwidth]{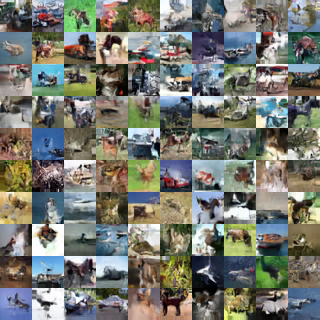}
        \text{(a) Cifar10}
        \label{fig:Cifar10}
    \end{minipage}
    \hfill{}
    \begin{minipage}[t]{0.31\columnwidth}
        \centering
        \begin{center}
        \vspace{0.05cm}
        \resizebox{\columnwidth}{!}{%
        \begin{tabular}{ |c|c|c| } 
         \cline{2-3}
         \multicolumn{1}{c|}{}  & $\downarrow$ bpd & $\downarrow$ FID \\ 
         \hline
         Naive sampling & 1.03 & 15.82 \\ 
         NF-SAILS & \textbf{0.96} & 14.12 \\ 
         WGAN-GP & N/A & {\textbf{12.89}} \\
         \hline
        \end{tabular}
        }
        \end{center}
        \includegraphics[width=\textwidth]{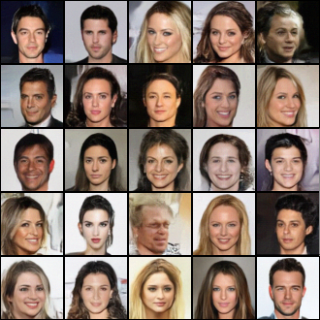}
        \text{(b) CelebA}
        \label{fig:CelebA}
    \end{minipage}
    \hfill
    \begin{minipage}[t]{0.31\columnwidth}
         \centering
        \begin{center}
        \resizebox{\columnwidth}{!}{%
        \begin{tabular}{ |c|c|c| } 
         \cline{2-3}
         \multicolumn{1}{c|}{}   & $\downarrow$ bpd & $\downarrow$ FID \\ 
         \hline
         Naive sampling & 2.38 & 8.91 \\ 
         NF-SAILS & \textbf{2.11} & \textbf{7.86} \\ 
         WGAN-GP & N/A &  9.56 \\
         \hline
        \end{tabular}
        }
        \end{center}
        \includegraphics[width=\textwidth]{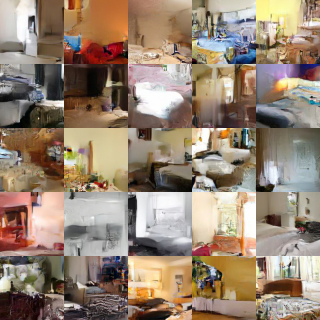}
        \text{(c) LSUN (bedroom)}
        \label{fig:LSUN}
    \end{minipage}
    \caption{Tables report quantitative and perceptual metrics computed from the samples generated by the compared methods. The figures show some samples generated from Glow using the proposed NF-SAILS method.}
    \label{fig:images}
\end{figure*}

We compare the FID score as well as the average negative log-likelihood (bpd), keeping all training conditions constant and averaging the results over 10 Monte Carlo runs. The results are depicted in Fig. \ref{fig:images} reports the results when compared to those obtained by naive sampling or WGAN-GP \citep{gulrajani2017improved}. As shown by the different panels of this figure, the proposed NF-SAILS method considerably improves the quality of the generated images, both quantitatively (in term of FID) and semantically. Our methodology compares favourably w.r.t. to WGAN-GP for the two data sets CelebA and LSUN.

\section{Conclusion}

This paper discusses the sampling from the target distribution learnt by a normalizing flow.  Architectural constraints prevent normalizing flows to properly learn disconnect support measures due to the topological mismatch between the latent and target spaces. Moreover, we theoretically prove that Jacobian norm of the transformation become arbitrarily large to closely represent such target measures. The conducted analysis exhibits the existence of pathological areas in the latent space corresponding to points with exploding Jacobian norms. Using a naive sampling strategy leads to out of distribution samples located in these areas. To overcome this issue, we propose a new sampling procedure based on a Langevin diffusion directly formulated in the latent space. This sampling is interpreted as a Riemanian manifold Metropolis adjusted Langevin algorithm, whose metrics is driven by the Jacobian of the learnt transformation. This local exploration of the latent space is complemented by an independent Metropolis-Hastings kernel which allows moves from one high probability region to another while avoiding crossing pathological areas. One particular advantage of the proposed is that it can be applied to any pre-trained NF model. Indeed it does not require a particular training strategy of the NF or to adapt the distribution assumed in the latent space. The performances of the proposed sampling strategy show to compare favorably to state-of-the art, with very few out-of-distribution samples. 

\begin{appendices}

\section{Proof of Proposition \ref{prop:jacobian}}
\label{sec:proof}

The proof of Proposition \ref{prop:jacobian} in Section \ref{subsec:topology_preservation} combines existing results from topology and real analysis. The complete background can be found in \citep{dudley2002real} and \citep{cornish2020relaxing}. The proof is mainly based on the following results.


\begin{theorem}[\citet{cornish2020relaxing}]
Let $q_Z$ and $q_X$ define probability measures on $\mathbb{R}^d$, with $\operatorname{supp}(q_Z) \not \neq $ $\operatorname{supp}(q_X)$. For any sequence of measurable, differentiable Lipschitzian functions $f_n: \ \mathbb{R}^{d} \rightarrow \mathbb{R}^{d}$, if the sequence weakly converges as $f_{n\#} q_Z \stackrel{\mathcal{D}}{\rightarrow} q_X$, then 
\begin{equation}
\lim _{n \rightarrow \infty} \operatorname{Lip} f_n=\infty.
\end{equation}
\label{theo:bilip}
\end{theorem}
Moreover, \citet{behrmann2019invertibility}  showed the relation between the Lipschitz constant and the Jacobian of a transformation, as stated below.
\begin{lemma}[Rademacher's theorem]
If $f: \mathbb{R}^m \rightarrow \mathbb{R}^n$  is Lipschitzian, then $f$ is continuous and differentiable at almost all points of $\mathbb{R}^m$ and
\begin{equation}
\operatorname{Lip} f= \sup _{z \in \mathcal{Z}}\| J_f(z)\|_{\mathrm{op}}
\end{equation}  
\label{lem:rademacher}
\end{lemma}
Both Theorem \ref{theo:bilip} and Lemma \ref{lem:rademacher} rely on the same starting hypothesis, i.e., $f$ is required to be continuous, differentiable and Lipschitzian. Combining these two results yields Proposition \ref{prop:jacobian} following a development of the proof of the results by \citet{cornish2020relaxing} and \citet{behrmann2019invertibility}. \\

\section{Properties of the Jacobian of coupling layer-based NFs} \label{app:sec:jacobian}
\subsection{Structure of the Jacobian matrix and computation of its determinant} \label{app:subsec:jacobian_structure}

RealNVP model defines a NF by implementing a sequence of $M$ invertible bijective transformation functions, herein referred to as coupling layers \citep{dudley2002real}. In other words, the mapping $f$ writes as $f = f^{(M)} \circ f^{(M-1)} \circ f^{(2)} \circ f^{(1)} $. Each bijection $f^{(m)}: u \mapsto v$ associated to the $m$th layer splits the input $u\in \mathbb{R}^D$ into two parts of sizes $d$ and $d-D$ ($d\leq D$), respectively, such that the output $v\in \mathbb{R}^D$ writes
\begin{equation}
\left\{
\begin{array}{ccl}
    v_{1: d} &=&u_{1: d} \\
v_{d+1: D} &=&u_{d+1: D} \odot \exp \left(h^{(m)}\left(u_{1: d}\right)\right)+{t}^{(m)}\left(u_{1: d}\right)
\end{array}
\right.
\end{equation}
where $h_m(\cdot): \mathbb{R}^d \rightarrow \mathbb{R}^{D-d}$ and $t_m(\cdot) :\mathbb{R}^d \rightarrow \mathbb{R}^{D-d}$ are scale and translation functions implemented as deep networks and $\odot$ stands for the Hadamard product. The Jacobian of the above transformation is a lower triangular matrix
\begin{equation}
J^{(m)}(u)=\left[\begin{array}{cc}
{I}_d & \boldsymbol{0}_{d \times(D-d)} \\
{A}^{(m)}(u)   & {E}^{(m)}(u)
\end{array}\right]    
\label{eq:rnvp_jac}
\end{equation}
where ${I}_d$ and $\boldsymbol{0}_{d \times(D-d)}$ are the identity  and  zero matrices with indexed sizes, respectively, and
\begin{equation}
\left\{
\begin{array}{ccl}
{A}^{(m)}(u) &=& u_{d+1: D} \odot \frac{\partial \exp h^{(m)}(u_{1: d})}{\partial u_{1: d}} + \frac{\partial t^{(m)}(u_{1: d})}{\partial u_{1: d}}  \\ 
{E}^{(m)}(u) &=&\operatorname{diag}\left(\exp \left(h^{(m)}\left(u_{1: d}\right)\right)\right).
\end{array}
\right.
\end{equation}
Thanks to the chain rule, it follows that the Jacobian of the overall NF is
\begin{equation} \label{eq:jacobian_formula}
    J_f(z) = \prod_{j=1}^J J^{(m)}(u^{(m)})
\end{equation}
with $u^{(m)} = f^{(m-1)}(u^{(m-1)})$ and $z = u^{(0)}$. 

Moreover, because of the structure of each layer, the determinant of the Jacobian $J^{(m)}(u)$ associated with the $m$th layer is
\begin{equation} \label{eq:_layer_determinan_formula}
    |J^{(m)}(u)| = \prod_{k=1}^d \exp \left(h^{(m)}\left(u_{k}\right)\right).
\end{equation}
The determinant of the Jacobian $J_f(\cdot)$ characterizing the overall NF can be easily computed from \eqref{eq:jacobian_formula} and \eqref{eq:_layer_determinan_formula}.

\subsection{Positive definiteness of the Jacobian}

\begin{property} \label{prop:psd}
    The product of two lower triangular matrices with strictly positive diagonal elements is a positive definite lower triangular matrix.
\end{property}

\renewcommand{\qedsymbol}{$\blacksquare$}
\begin{proof} 
Let $A=\left[a_{i j}\right]$ and $B=\left[b_{i j}\right]$ be two $n \times n$ lower triangular matrices with positive diagonal entries, i.e., 
\begin{eqnarray}
     \forall i,j& \ \text{such that} \ i<j,\ \text{then}\  a_{i j}=b_{i j}=0. & \\
      & \forall i \  a_{ii}>0 \ \text{and} \ b_{i i}>0 &
\end{eqnarray}
Let $C=\left[c_{i j}\right]$ denote the  product matrix $C=A B$ with $c_{i j}=\sum_{k=1}^n a_{i k} b_{k j}$. The upper elements $c_{i j}$ ($i<j$) of $C$ can be computed as
\begin{equation} \label{eq:matrix_C_elements}
    c_{i j} = \sum_{k=1}^i a_{i k} b_{k j}+\sum_{k=i+1}^n a_{i k} b_{k j}.
\end{equation}
In the right hand side of \eqref{eq:matrix_C_elements}, if $k \leq i$ then $b_{kj=0}$. Moreover if $k> i$ then $a_{ik}=0$. As a consequence, $c_{ij}=0$ and $C$ is triangular.

Moreover, the eigenvalues of a triangular matrix is its diagonal elements. It follows that $C$ is positive definite.
\end{proof}

Thanks to the structure of coupling layer-based NFs discussed in Appendix \ref{app:subsec:jacobian_structure}, we have the two following corollaries.

\begin{corollary}
   The Jacobian matrix $J_f(\cdot)$ and its inverse $J_f^{-1}(\cdot)$ of coupling layer-based NFs are positive definite.
\end{corollary}

\begin{corollary}
   The matrix $G(\cdot)$ and its inverse $G^{-1}(\cdot)$ are positive definite.
\end{corollary}

\section{Diffusion in the latent space}
\label{sec:lattent_diff}

\subsection{Preliminaries}\label{app:subsec:preliminaries}

The Langevin diffusion is a particular instance of the Itô process defined in the following Lemma of which a proof is given in  \citep{oksendal2003stochastic}.

\begin{lemma}[Itô's lemma]
Let $X_t$ denote an Itô drift-diffusion process defined by the stochastic differential equation
\begin{equation} \label{eq:ito_diffusion}
    d X_t=\mu_t d t+\sigma_t d B_t.
\end{equation}
If $f: \mathbb{R}^2 \rightarrow \mathbb{R}$  is a differentiable scalar function, then
\begin{equation}
    d f\left(t, X_t\right)=\left(\frac{\partial f}{\partial t}+\mu_t \frac{\partial f}{\partial x}+\frac{\sigma_t^2}{2} \frac{\partial^2 f}{\partial x^2}\right) d t+\sigma_t \frac{\partial f}{\partial x} d B_t.
    \label{ito}
\end{equation}
It yields that $f\left(t, X_t\right)$ is an Itô drift-diffusion process itself.
\label{lem:ito}

\end{lemma}




The following property shows that for any bijective transformation, the Jacobian of the inverse transformation is equal to the inverse of the Jacobian of the transformation. This result will be useful later. 

\begin{property} \label{prop:jaco_trick}
Let $f: \mathcal{Z} \rightarrow \mathcal{X}$ denote a bijective transformation  and $J_f(\cdot)$ its Jacobian, then 
$$J_{f^{-1}}(f(z))=J_{f}^{-1}(z).$$   
\end{property}

\begin{proof}
Let $h$ and $g$ denote two multivariate functions. The chain rule writes
\begin{equation}
    J_{h \circ g}(\cdot)=J_h(g(\cdot)) J_g(\cdot)
\end{equation}
thus
\begin{equation}\label{eq:chain_rule}
    J_h(g(\cdot)) = J_{h \circ g}(\cdot) J_g^{-1}(\cdot).
\end{equation}
Moreover, for any multivariate bijective function $f$, we have 
\begin{equation} \label{eq:prop_jacobian}
   J_{f \circ f^{-1}}(\cdot)=J_{f^{-1} \circ f}(\cdot)={I}_d.
\end{equation}
Combining \eqref{eq:chain_rule} and \eqref{eq:prop_jacobian} with $h=f^{-1}$ and $g=f$ yields
\begin{equation}
J_{f^{-1}}(f(z))=J_{f^{-1} \circ f} (z) J_{f}^{-1}(z)= {I}_d J_{f}^{-1}(z) = J_{f}^{-1}(z).
\end{equation} 
\end{proof}

The following property demonstrates that the gradient of the score of $q_X$ can be expressed over the latent space $\mathcal{Z}$ using $\tilde{q}_Z$ defined in \eqref{eq:q_tilde}.

\begin{property} \label{prop:score}
    Let $f: \mathcal{Z} \rightarrow \mathcal{X}$ be a bijective transformation which maps a latent measure $q_Z$ towards a target measure $q_X$. Then the score of $q_X(x)$ is given by
    $$
    \nabla_{x} \log {q}_X(x) = J_{f}^{-1}(z) \cdot \nabla_{z} \log \tilde{q}_Z(z)  
    $$
    where $\tilde{q}_Z(z) = q_Z(z) \left|J_{f}(z)\right|^{-1}$.
\end{property}

\begin{proof} 
From the definition of ${q}_X(x)$ in equation \eqref{eq:change_of_variable}, the score of $q_X(x)$  writes
\begin{equation}
    \nabla_{x} \log {q}_X(x) = \nabla_{x} \left[ \log q_Z(f^{-1}(x))+\log |J_{f^{-1}}(x)| \right]
\end{equation} 
and, from Property \ref{prop:jaco_trick},
\begin{align}
    \nabla_{x} \log {q}_X(x) & = \nabla_{f(z)} \left[\log q_Z(z) + \log \left|J_{f}(z)\right|^{-1} \right] \\
     & = \nabla_{f(z)} \log \tilde{q}_Z(z) \label{eq:q_z_x}
\end{align} 
The chain rule now leads to
\begin{align}
    \nabla_{x} \log {q}_X(z) & = \nabla_{x} f^{-1}(x) \cdot \nabla_{z} \log \tilde{q}_Z(z) \\
    & = J_{f^{-1}}(f(z)) \cdot \nabla_{z} \log \tilde{q}_Z(z)
\end{align} 
which, using Property \ref{prop:jaco_trick}, can be finally rewritten as
\begin{equation}
    \nabla_{x} \log {q}_X(x) = J_{f}^{-1}(z) \cdot \nabla_{z} \log \tilde{q}_Z(z).
\end{equation}
\end{proof}

\subsection{Derivation of the proposal distribution}\label{subsec:appendix_derivation_proposal}

The following property shows that the Lanvegin diffusion which targets the distribution $q_X$ can be rewritten as a diffusion over the latent space $\mathcal{Z}$.

\begin{property} \label{prop:latentdiff}
   We consider the overdamped Langevin Itô diffusion
\begin{equation}
dX_t = \nabla_{x} \log q_X(X_t) dt + \sigma_t d B_t    
\label{MALA}
\end{equation}
driven by the time derivative of a standard Brownian motion $B_t$. In the limit $t \rightarrow \infty$, this probability distribution $X_t$ approaches a stationary distribution $q_X$.  Let $f: \mathcal{Z} \rightarrow \mathcal{X}$ be a bijective transformation which maps a latent measure $q_Z$ towards the target measure $q_X$. A counterpart Langevin diffusion expressed over the latent space $\mathcal{Z}$ writes
\begin{equation}
dZ_{t} = G^{-1}(Z_t) \nabla_{z} \log \tilde{q}_Z(Z_t) dt +  \sigma_t \sqrt{G^{-1}(Z_t)} d B_t   
\end{equation}
\end{property}

\begin{proof} 
The Langevin diffusion is a particular instance of the Itô process where the drift $\mu_t$ in \eqref{eq:ito_diffusion} is given by the gradient of the log-density $\nabla_{x} \log q_X(X_t)$, i.e., 
\begin{equation}
dX_t = \nabla_{x} \log q_X(X_t) dt + \sigma_t d B_t    
\end{equation}

We are interested in the diffusion process of $f^{-1}(X_t)$ when $f(\cdot)$ is a NF which is continuous, differentiable and bijective such that $f(Z_t)=X_t$ and $f^{-1}(X_t)=Z_t$. The Îto's Lemma \ref{lem:ito} states
\begin{equation}
d f^{-1}(X_t) = \left( J_{{f}^{-1}}(X_t) \nabla_{x} \log q_X(X_t) + \frac{\sigma_t^2}{2} \mathrm{tr}(H_{{f}^{-1}}(X_t)) \right) d t+ \sigma_t J_{f^{-1}}(X_t) d B_t.
\label{simpim}
\end{equation} 
Neglecting the second-order terms yields
\begin{equation}
  df^{-1}(X_{t}) = J_{f^{-1}}(X_t) \nabla_{x} \log q_X(X_t) dt+ \sigma_t J_{{f}^{-1}}(X_t) d B_t. 
\label{diffchange}
\end{equation}
Using Property \ref{prop:jaco_trick}, Eq. \eqref{diffchange} can be rewritten as
\begin{equation}
  dZ_t =J_{f}^{-1}(Z_t)  \nabla_{x} \log q_X(X_t) dt + \sigma_t J_{{f}}^{-1}(Z_t) d B_t.
\label{diffchange2}
\end{equation}
Finally, by denoting $G^{-1}(z) = \left[J_{f}^{-1}(z)\right]^2$ and using Property \ref{prop:score}, the diffusion in the latent space writes
\begin{equation}
\label{eq:RiDiff}
dZ_{t} = G^{-1}(Z_t) \nabla_{z} \log \tilde{q}_Z(Z_t) dt +  \sigma_t \sqrt{G^{-1}(Z_t)} d B_t   
\end{equation} 
\end{proof}

The discretization of the stochastic differential equation \eqref{eq:RiDiff} using the Euler–Maruyama scheme can be written as in \eqref{eq:latentlangevin}. This discretized counterpart of the diffusion corresponds to the proposal move of a Riemann manifold Metropolis-Adjusted Langevin algorithm which targets $\tilde{q}_Z$. The following property shows that the associated proposal kernel can be rewritten as \eqref{proposal}.


\begin{property}\label{prop:kernel}
    The discrete Langevin diffusion given by 
\begin{equation}
  z^{\prime} = z + \frac{\epsilon^2}{2} \cdot G^{-1}(z) \nabla_{z} \log \tilde{q}_Z(z) + \epsilon \cdot \sqrt{G^{-1}(z)} \xi 
\label{discretechange}
\end{equation}    
        with $\xi \sim \mathcal{N}(0, I)$ is defined by the transition kernel
    \begin{equation}
    q\left(z^{\prime} \mid z\right) \propto |J_{f}(z)| \exp\left[- \frac{1}{2\epsilon^{2}} \left \| J_{f}(z) (z^{\prime}-z) + \frac{\epsilon^2}{2} J_{f}^{-1}(z)  \nabla_{z} \log \tilde{q}(z) \right \|^2 \right].
    \label{eq:proposal_append}
    \end{equation}

\end{property}

\begin{proof}
From the Gaussian nature of $\xi$, the conditional distribution of $z^{\prime}$ is a Gaussian distribution whose mean is governed by the drift and covariance matrix is parametrized by the (inverse of) the Jacobian, namely
\begin{equation}
    z^{\prime} \mid z \sim \mathcal{N}(\mu,\Sigma)
\end{equation}
with
\begin{align}
    \mu &= z + \frac{\epsilon^2}{2} \cdot G^{-1}(z) \nabla_{z} \log \tilde{q}_Z(z) \label{eq:mu}\\
    \Sigma &= \epsilon^2  J_{f}^{-1}(z) J_{f}^{{-\top}}. \label{eq:Sigma}
\end{align}
The corresponding pdf writes
\begin{equation}\label{eq:proposal_canonical}
    q\left(z^{\prime} \mid z\right)  = \left(\frac{1}{2 \pi}\right)^{\frac{d}{2}} \frac{1}{|\Sigma|^{1/2}}  \exp \left(-\frac{1}{2}(z'-\mu)^{\top}\Sigma^{-1}(z'-\mu)\right).
\end{equation}
First, let notice that $\Sigma^{-1} = \epsilon^{-2} J_{f}^{\top}(z) J_{f}(z)$. Then we have
\begin{align}
    (z'-\mu)^{\top}\Sigma^{-1}(z'-\mu) &= \epsilon^{-2} \left[z' - z - \frac{\epsilon^2}{2} \cdot \left[J_f^{-1}(z)\right]^2 \nabla_{z} \log \tilde{q}_Z(z) \right]^{\top}  J_{f}^{\top}(z) \\
     & \qquad \times J_{f}(z)\left[z' - z - \frac{\epsilon^2}{2} \cdot \left[J_f^{-1}(z)\right]^2 \nabla_{z} \log \tilde{q}_Z(z)\right] \nonumber\\
     & = \epsilon^{-2} \left \| J_{f}(z)(z^{\prime}-z) - \frac{\epsilon^2}{2} J_{f}^{-1}(z) \nabla_{z} \log \tilde{q}(z) \right \|^2
\end{align}
Finally, using $|\Sigma|^{1/2} = \epsilon |J_{f}^{-1}(z)| =  \epsilon {|J_{f}(z)|^{-1}}$ yields
\begin{equation}
\label{eq:proposal_final}
    q\left(z^{\prime} \mid z\right) \propto |J_{f}(z)| \exp\left[- \frac{1}{2\epsilon^{2}} \left \| J_{f}(z) (z^{\prime}-z) - \frac{\epsilon^2}{2} J_{f}^{-1}(z) \nabla_{z} \log \tilde{q}(z) \right \|^2\right].
\end{equation}
\end{proof}

It is worth noting that the pdf of this transition kernel should be computed when evaluating the acceptance ratio \eqref{eq:MALA-alp}. When using the canonical writing \eqref{eq:proposal_canonical}, evaluating this pdf would require to compute $G^{-1}(z)$ in \eqref{eq:mu} and $J_{f}^{-1}(z) J_{f}^{{-\top}}(z)$ in  \eqref{eq:Sigma}. Instead, evaluating this pdf with the specific form \eqref{eq:proposal_final} only requires to compute $J_{f}(z)$ since the latent score $\tilde{s}_Z(z)=J_{f}^{-1}(z) \nabla_{z} \log \tilde{q}(z)$ can be computed efficiently, as discussed later in Appendix \ref{app:subsubsec:fast_comput}.


\subsection{Efficient implementation}\label{app:subsec:efficient_implem}
\subsubsection{Approximation of the proposal move}\label{app:subsubsec:approximation_expansion}
Generating  high dimension Gaussian variables according to \eqref{eq:latentlangevin} is expected to be very costly because of the covariance matrix, even if the corresponding Cholesky factor $\epsilon J_f^{-1}(\cdot)$ is lower triangular and explicit (see Appendix \ref{app:subsec:jacobian_structure}). Alternatively, to lighten the computation, we take advantage of the $1$st order expansion
\begin{equation}
    f^{-1}\left(f(z_k) +\epsilon \xi \right) \simeq f^{-1} \circ f(z_k) + \epsilon J_{f^{-1}}(f(z_k))  \xi.
\end{equation}
Using Property \ref{prop:jaco_trick}, this amounts to approximate \eqref{eq:latentlangevin} by the diffusion
\begin{eqnarray}
z^{\prime} & = f^{-1}\left(f(z_k) +\epsilon \xi \right) + \frac{\epsilon^2}{2} G^{-1}(z_k) \nabla_{z} \log \tilde{q}(z) \\
& = f^{-1}\left(f(z_k) +\epsilon \xi \right) + \frac{\epsilon^2}{2} J_f^{-1}(z_k) \ts_Z(z).
\label{eq:difusion}
\end{eqnarray}
According to \eqref{eq:difusion}, this alternative proposal scheme only requires to generate high dimensional Gaussian variables whose covariance matrix is now identity.

\subsubsection{Fast computation of the latent score} \label{app:subsubsec:fast_comput}

The latent score $\ts_Z(z)$ is a critical quantity in the proposed method, as it contributes to the drift term in the proposal move \eqref{eq:diffusion_eff} and to the  proposal kernel \eqref{proposal}. Property \ref{prop:score} shows that the latent score is equal to the score of $q_X$ expressed in the target domain, i.e.,
$$
\nabla_{x} \log {q}_X(x) = J_{f}^{-1}(z) \cdot \nabla_{z} \log \tilde{q}_Z(z)  
$$ 
The adopted implementation bypasses the costly evaluation and storage of the inverse Jacobian by directly computing the latent score as $\nabla_{x} \log {q}_X(x)$. The evaluation of the score of $q_X$ can be conveniently performed thanks to the auto-differentiation modules provided by numerous deep learning frameworks.

\section{Experiments}\label{app:sec:experiments}

\subsection{Training}\label{app:subsec:training}
In all experiments we trained our models to maximize either the log-likelihood using the ADAM optimiser with default hyperparameters and no weight decay. We used a held-out validation set and trained each model until its validation score stopped improving, except for the synthetic data experiments where we train for a fixed number of 1000 epochs.

\subsection{Complementary results for the synthetic experiments}\label{app:sec:complementary}

\begin{figure}
\centering
\includegraphics[width=0.24\linewidth]{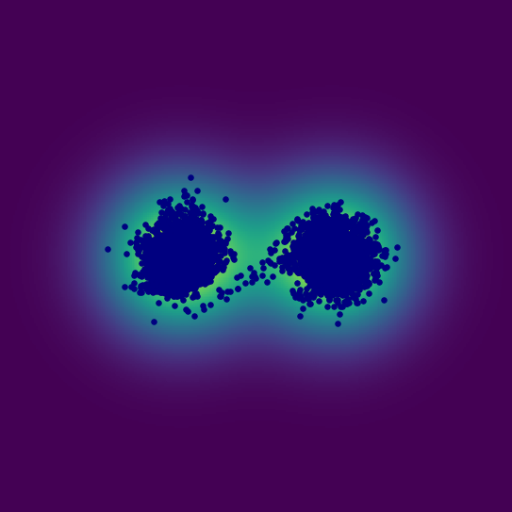}\includegraphics[width=0.24\linewidth]{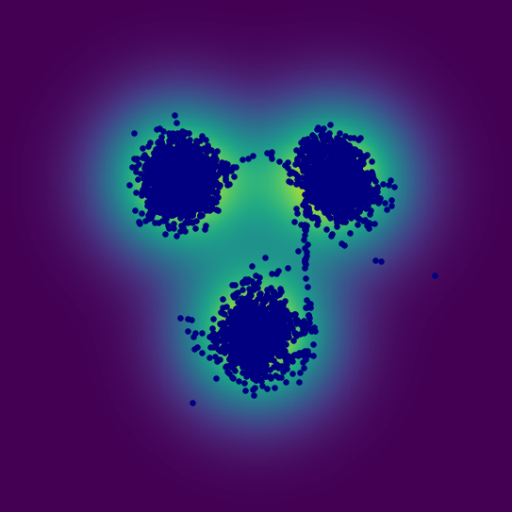}\includegraphics[width=0.24\linewidth]{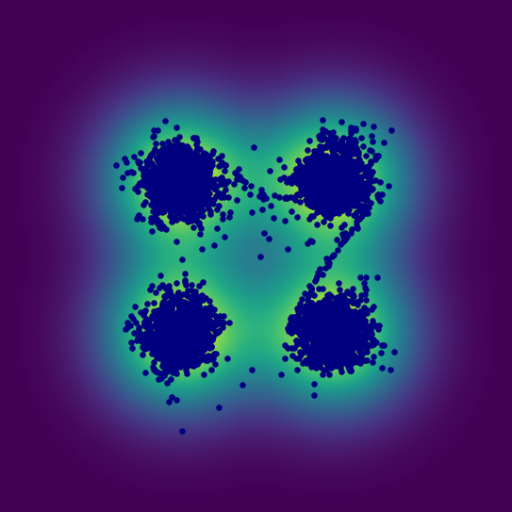}\includegraphics[width=0.24\linewidth]{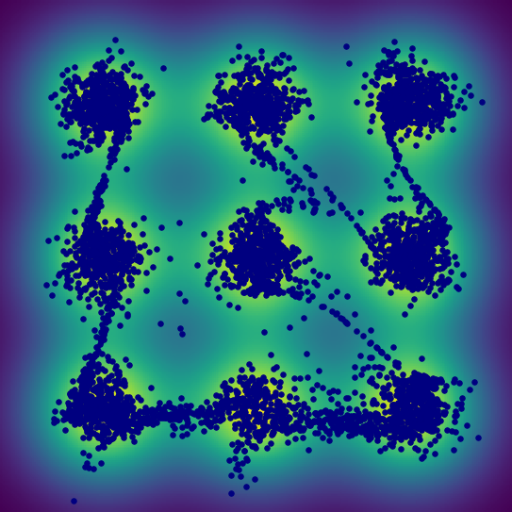}\\[-0.9pt]
\includegraphics[width=0.24\linewidth]{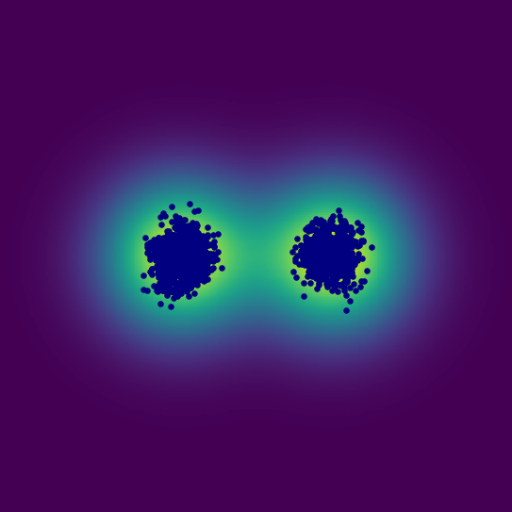}\includegraphics[width=0.24\linewidth]
{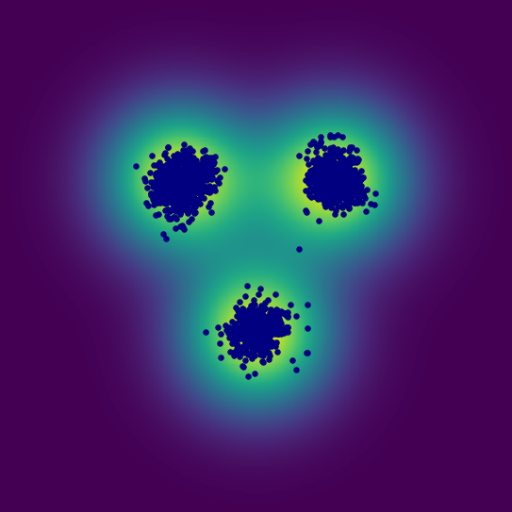}\includegraphics[width=0.24\linewidth]{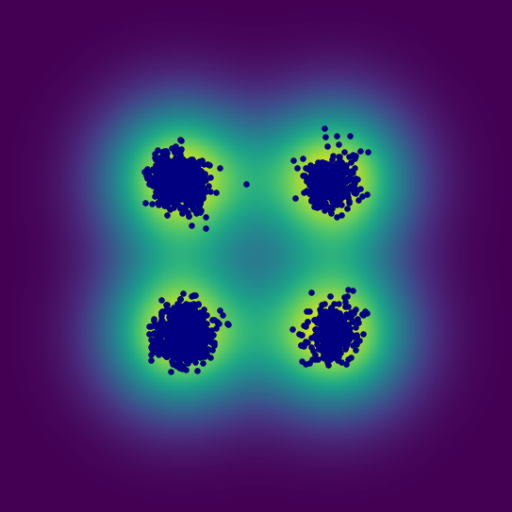}\includegraphics[width=0.24\linewidth]{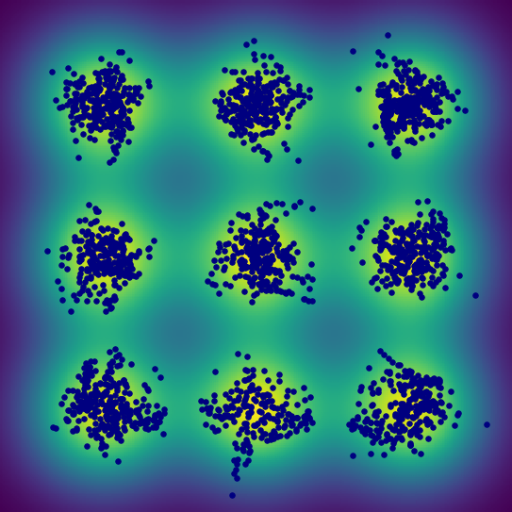}
\caption{{Mixture of $k$ Gaussian distributions (green), and 1000 samples (blue) generated by the naive sampling (top) and the proposed NF-SAILS method (bottom) with, from left to right, $k=2$, $k=3$, $k=4$ and $k=9$.} \label{fig:kGaussians}}
\end{figure}

Figure \ref{fig:kGaussians} shows the difference of sampling quality between naive sampling and the proposed NF-SAILS method for RealNVP model trained on $k$-mixtures of Gaussians for $k \in \left\{2,3,4,9\right\}$. See also Table \ref{tab:Gaussians} in Section \ref{subsec:results_synth} of the main document.

\subsection{Implementation details for the image experiments}\label{app:sec:implementation_details}

The hyperparameters used in the experiments conducted on images (see Section 6.3 of the main document) are reported in Table \ref{tab:hyper}.

\begin{table}[!ht]
    \centering
    \begin{tabular}{|l|c|c|c|c|}
    \hline
        Dataset & Minibatch Size & Levels (L) & Depth per level (K) & Coupling \\ \hline \hline
        CIFAR-10 & 512 & 3 & 32 & Affine \\ \hline
        LSUN, $64\times64$ & 128 & 4 & 48 & Affine \\ \hline
        LSUN, $96\times96$ & 320 & 5 & 64 & Affine \\ \hline
        LSUN, $128\times128$ & 160 & 5 & 64 & Affine \\ \hline
        CelebA, $96\times96$ & 320 & 5 & 64 & Affine \\ \hline
        CelebA, $128\times128$ & 160 & 6 & 32 & Affine \\ \hline
    \end{tabular}
    \caption{Architectures of the Glow model implemented for the experiments conducted on the  image data sets.}\label{tab:hyper}
\end{table}

\end{appendices}

\bibliographystyle{abbrvnat}
\bibliography{strings_all_ref,reference}
\end{document}